\documentclass[11pt]{article}

\usepackage[preprint]{acl}

\usepackage{times}
\usepackage{latexsym}

\usepackage[T1]{fontenc}

\usepackage[utf8]{inputenc}

\usepackage{microtype}

\usepackage{inconsolata}

\usepackage{graphicx}

\usepackage{amsmath}
\usepackage{amssymb}
\usepackage{mathtools}
\usepackage{amsthm}
\usepackage{subcaption}
\usepackage{booktabs} 
\usepackage{multirow}
\usepackage{colortbl}
\usepackage{makecell}

\title{GAMMA: Global Bit Allocation for Mixed-Precision Models under Arbitrary Budgets}

\author{
 \textbf{Zhangyang Yao\textsuperscript{1}},
 \textbf{Haiyan Zhao\textsuperscript{2}},
 \textbf{Haoyu Wang\textsuperscript{2}},
 \textbf{Xu Han\textsuperscript{2}}
\\
 \textsuperscript{1}Beihang University,
 \textsuperscript{2}Tsinghua University,
\\
}

\newcommand{\MethodName}{\textsc{GAMMA}}

\begin{document}
\maketitle
\begin{abstract}
Mixed-precision quantization improves the budget--accuracy trade-off for large language models (LLMs) by allocating more bits to sensitive modules.
However, automating this allocation at LLM scale faces a unique combination of constraints: learnable approaches require quantization-aware training, which is \textbf{infeasible for billion-parameter models}; training-free alternatives rely on static proxy metrics that \textbf{miss cross-module interactions} and must be recomputed per target budget; and search-based methods are expensive \textbf{without guaranteeing exact budget compliance}.
We propose \MethodName, a quantizer-agnostic framework that learns module-wise precision preferences \textbf{entirely within a post-training pipeline}.
\MethodName~ optimizes a teacher-forced hidden-state reconstruction objective under an augmented Lagrangian constraint, and projects the learned preferences into exact budget-feasible discrete assignments via integer programming.
A key property is \textbf{score reuse}: because the learned preferences encode a stable sensitivity ranking rather than budget-specific weights, a single training run serves arbitrary deployment targets by re-solving only the integer program, \textbf{reducing per-budget adaptation from hours to a few minutes}.
Across Llama and Qwen models (8B--32B), \MethodName~ outperforms both fixed-precision baselines (up to \textbf{+12.99} Avg.) and search-based mixed-precision methods (up to \textbf{+7.00} Avg.), and can match fixed 3-bit quality at \textbf{2.5-bit} average precision, enabling deployment at substantially smaller memory footprints.

\end{abstract}

\section{Introduction}

LLMs have improved rapidly as their scale grows, but deploying them remains constrained by memory footprint and memory bandwidth.
These constraints are especially severe in on-device and edge deployment, where even a single deployed model can exceed the available capacity or lead to unacceptable latency.
Post-training quantization (PTQ) is one of the most practical ways to reduce deployment cost, as it requires only a small calibration set and no full retraining.
In particular, weight-only PTQ---which compresses model weights to low bit-widths while keeping activations in higher precision---has become the dominant paradigm for LLM deployment.
While recent weight-only PTQ methods often preserve most of the original accuracy at 4 bits, the regime changes below 4 bits.
In the extremely low-bit setting, uniform-bit quantization can exhibit sharp drops in capability, and small changes in precision can lead to disproportionate accuracy loss.

A natural way to operate under diverse memory budgets is mixed-precision quantization.
Instead of quantizing all modules to the same bit-width, mixed precision allocates higher precision to sensitive components and lower precision elsewhere, achieving non-integer average bit-widths (e.g., 2.5 or 3.5 bits) and improving the budget--accuracy trade-off.
The central difficulty is bit allocation: deciding where limited bits should be spent.
Existing approaches typically rely on (i) search over a large discrete space of assignments, or (ii) handcrafted proxy objectives to estimate module importance.
Both paradigms face fundamental limitations in the LLM setting.
Search-based allocation is expensive at scale and cannot guarantee exact budget compliance.
Hessian-based sensitivity proxies rely on static per-layer metrics that may not capture cross-module dependencies and require recomputation for each target budget.
More critically, most mixed-precision methods with learnable allocation---including differentiable approaches and sparsity-based regularization---require quantization-aware training (QAT), which is prohibitively expensive for billion-parameter models.
\textbf{This leaves a gap: existing methods do not simultaneously learn module-wise bit preferences without retraining weights, enforce exact global bit-width budgets, and reuse learned preferences across deployment targets---all within a pure post-training pipeline}.

We propose \MethodName~, a quantizer-agnostic framework that \textbf{addresses these challenges within a pure PTQ pipeline}.
\MethodName~ treats pre-quantized weight candidates at multiple bit-widths as a candidate pool, \textbf{freezes all model weights}, and learns only lightweight bit-selection parameters.

\MethodName~ follows a two-stage pipeline designed around LLM-specific constraints.
In Stage~I, we learn budget-aware precision preferences for each linear module via a differentiable relaxation.
Since end-to-end task losses tend to overfit small calibration sets, we instead optimize a teacher-forced, layer-wise hidden-state reconstruction loss that captures how quantization error propagates through intermediate representations.
To enforce exact budget compliance, we formulate the global bit-width constraint via an augmented Lagrangian with learned dual variables~\cite{wang2020structured}, rather than a soft penalty that may leave the budget misaligned.
In Stage~II, we convert the learned soft preferences into an exact, budget-feasible discrete assignment using integer programming, enabling reproducible deployment under strict memory constraints.
A key practical advantage is that the learned preference scores can be \textbf{reused across budgets}: changing the target only requires re-solving the integer program, \textbf{reducing per-budget adaptation from hours to a few minutes} rather than retraining.
We validate \MethodName~ on both QTIP~\cite{tseng2024qtip} and AQLM~\cite{egiazarian2024extreme} backends, confirming that the allocation quality is \textbf{not tied to a particular quantizer} (Appendix~\ref{app:aqlm_backend}).

Empirically, \MethodName~ consistently outperforms both fixed-precision and search-based mixed-precision baselines across Llama and Qwen models (8B--32B).
Under extreme compression (2.5--3.0 bits), \MethodName~ improves over uniform quantization by up to \textbf{+12.99} Avg.\ and over search-based AMQ by up to \textbf{+7.00} Avg.
Under the same PTQ backbone, \MethodName~ matches fixed 3-bit quality at \textbf{2.5-bit} average precision, effectively pushing practical deployment toward lower-bit regimes without severe capability collapse.

In summary, \MethodName~ fills the gap identified above with three key properties: (i) it learns \textbf{module-wise sensitivity without weight retraining}, operating entirely within a post-training pipeline; (ii) it enforces \textbf{exact budget compliance} during differentiable optimization, avoiding the misalignment of soft penalties; and (iii) it produces \textbf{reusable allocation scores}---a single training run serves arbitrary deployment targets.

\section{Related Work}

\subsection{Post-Training Quantization for LLMs}
PTQ has become a primary approach for compressing LLMs due to its low computational cost and the ability to avoid full retraining.
PTQ methods can be broadly categorized into scalar quantization (e.g., GPTQ~\cite{frantar2022gptq}, AWQ~\cite{lin2024awq}, SmoothQuant~\cite{xiao2023smoothquant}, FlatQuant~\cite{sun2024flatquant}, OSTQuant~\cite{hu2025ostquant}) and vector quantization (e.g., AQLM~\cite{egiazarian2024extreme}, QuIP~\cite{chee2023quip}, QuIP\#~\cite{tseng2024quip}, QTIP~\cite{tseng2024qtip}). 
While these methods maintain high accuracy at 4 bits, uniform-bit quantization often exhibits sharp capability drops in the extremely low-bit regime.

\subsection{Mixed-Precision Quantization}
Mixed-precision quantization allocates higher precision to sensitive components to improve the budget-accuracy trade-off under strict global constraints (PB-LLM~\cite{shang2023pb}; BitStack~\cite{wang2024bitstack}). 
The central challenge is automated bit allocation across numerous modules.
Most prior methods with learnable allocation require QAT: DNAS~\cite{wu2018mixed} uses differentiable Gumbel-Softmax search but jointly trains weights with a task-level loss and a soft cost penalty, requiring full retraining after search; BSQ~\cite{yang2021bsq} and MSQ~\cite{han2025msq} exploit bit-level sparsity regularization, also under QAT. These methods were developed for CNNs and are difficult to scale to LLMs.
Training-free alternatives include HAWQ-V2~\cite{dong2020hawq}, which uses static Hessian-trace metrics for per-layer sensitivity ranking, and AMQ~\cite{lee2025amq}, which performs evolutionary search over discrete assignments for LLMs.
However, Hessian metrics do not capture cross-module interactions and require recomputation per target budget; search is expensive and must be repeated per budget.

\MethodName~ differs by learning allocation preferences via differentiable relaxation in a \textbf{pure PTQ setting}---\textbf{without weight retraining}---and enforcing \textbf{exact budget compliance} through constrained optimization.
The learned scores can be \textbf{reused across budgets} by re-solving the integer program.

\section{Method}

\subsection{Method Overview}

We aim to construct a mixed-precision quantized model under a prescribed average bit-width budget. 
Given a full-precision model and a set of pre-quantized weight candidates at multiple bit-widths, our goal is to assign a bit-width to each linear module such that the resulting mixed-precision model minimizes quantization-induced distortion while meeting the global memory budget.
A key design principle is that \textbf{all model weights remain frozen}; only lightweight bit-selection parameters are optimized.
This decoupling separates what the PTQ backbone quantizes (fixed) from how bits are allocated across modules (learned), and is what ultimately enables the resulting scores to transfer across deployment budgets.

\begin{figure*}[t] 
    \centering
        \includegraphics[width=\textwidth]{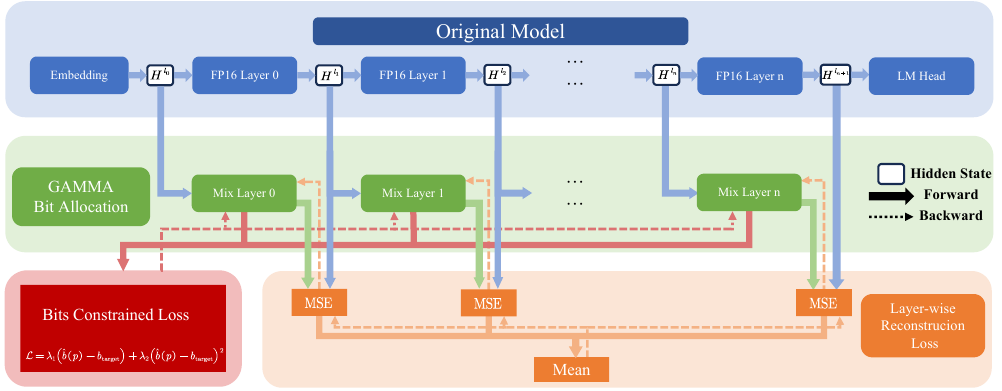}
\caption{\textbf{Training pipeline of \MethodName.}
Layer-wise hidden states from a full-precision teacher supervise mixed-precision mask learning via a reconstruction loss, while a global penalty enforces the target average bit-width.
Solid arrows indicate forward computation and dashed arrows indicate backward gradients.}
    \label{fig:train} 
\end{figure*}

Our approach consists of two stages.
\textbf{Stage I: Differentiable score learning.}
We introduce trainable multi-class masks that, for each linear module, produce a categorical distribution over candidate bit-widths.
Using continuous relaxation (Gumbel-Softmax), we optimize these masks on a calibration set to minimize a layer-wise hidden-state reconstruction objective while steering the \emph{expected} average bit-width toward a target budget via a constrained optimization objective.
\textbf{Stage II: Budget-feasible discretization.}
Since the learned masks are generally soft distributions rather than exact one-hot selections, we convert them into discrete bit-width assignments by solving a global constrained optimization problem.
This post-processing step selects exactly one bit-width per module to maximize consistency with the learned scores, subject to a global bit budget.
It can be implemented efficiently with knapsack-style dynamic programming or integer linear programming, following standard adaptive bit-width allocation formulations. Figure~\ref{fig:train} shows the training pipeline of \MethodName.

\subsection{Problem Setup}
\label{sec:problem_setup}

We consider a decoder-only Transformer with $L$ stacked layers.
We index layers by $i \in \{1,\ldots,L\}$ and linear modules within each layer by $j \in \mathcal{J}$, where
$\mathcal{J}=\{q,k,v,o,\text{up},\text{gate},\text{down}\}$ denotes the standard projection modules in self-attention and the feed-forward network.

Let $\mathcal{B}$ denote the set of candidate bit-widths supported by the underlying post-training quantization (PTQ) method.
For each module $(i,j)$ and each $b \in \mathcal{B}$, we assume a corresponding quantized weight tensor $W^{(i,j)}_{b}$ is available, produced by an off-the-shelf PTQ procedure.
We denote the full-precision weights by $W^{(i,j)}_{\mathrm{fp}}$; they are used only to define reference activations.

Our goal is to assign exactly one bit-width to each module.
We introduce binary assignment variables
\[
z_{i,j,b} \in \{0,1\}, \qquad \sum_{b \in \mathcal{B}} z_{i,j,b} = 1,
\]
and define the mixed-precision weight of module $(i,j)$ as
\begin{equation}
W^{(i,j)}_{\mathrm{mix}} = \sum_{b \in \mathcal{B}} z_{i,j,b}\, W^{(i,j)}_{b}.
\end{equation}

Let $N_{i,j}$ denote the number of parameters in module $(i,j)$.
The average bit-width of a mixed-precision assignment $z$ is
\begin{equation}
\bar{b}(z) =
\frac{
\sum_{i=1}^{L}\sum_{j \in \mathcal{J}} N_{i,j} \sum_{b \in \mathcal{B}} b \cdot z_{i,j,b}
}{
\sum_{i=1}^{L}\sum_{j \in \mathcal{J}} N_{i,j}
}.
\end{equation}

Given a target average bit-width $b_{\text{target}}$, we seek an assignment $z$ that minimizes quantization-induced distortion while meeting the global budget.
In Stage~I, we optimize a continuous relaxation of $z$ using categorical probabilities $p_{i,j,b}\in[0,1]$ with $\sum_{b}p_{i,j,b}=1$; Stage~II then projects the learned soft assignments back to a feasible binary solution $z$ under the target budget.

\subsection{Objective: Hidden-State Reconstruction}

Optimizing task-level accuracy directly is impractical in the post-training setting: calibration data are limited and bit-width assignments are discrete.
We therefore adopt a representation-level surrogate objective that matches the internal hidden states of the mixed-precision model to those of the full-precision reference.
Let $f_i(\cdot; W^{(i,*)})$ denote the computation of the $i$-th decoder layer parameterized by its collection of linear modules.
Given an input $x$ from the calibration set, let $H^{(l_i)}(x)$ be the hidden representation produced by the full-precision model after layer $i$:
\begin{equation}
H^{(l_i)}(x) = f_i\!\left(H^{(l_{i-1})}(x); W^{(i,*)}_{\mathrm{fp}}\right).
\end{equation}

To isolate per-layer quantization effects and stabilize optimization, we use a teacher-forced layer-wise scheme.
Specifically, for each layer $i$, we feed the full-precision hidden state $H^{(l_{i-1})}(x)$ as input to the mixed-precision version of layer $i$, yielding
\begin{equation}
H^{(l_i)}_{q}(x) = f_i\!\left(H^{(l_{i-1})}(x); W^{(i,*)}_{\mathrm{mix}}\right).
\end{equation}
We then minimize the layer-wise reconstruction error on the calibration distribution $\mathcal{D}_{\mathrm{cal}}$:
\begin{equation}
\ell_{\mathrm{mse}}
=
\frac{1}{L}\sum_{i=1}^{L}\;
\mathbb{E}_{x \sim \mathcal{D}_{\mathrm{cal}}}
\left[
\left\| H^{(l_i)}_{q}(x) - H^{(l_i)}(x) \right\|_2^2
\right].
\end{equation}
This objective directly measures quantization-induced distortion at each layer and serves as a \textbf{stable proxy for preserving the behavior of the full-precision model}.
We empirically confirm that replacing this reconstruction objective with a standard cross-entropy loss on the final output degrades performance by up to $-3.29$ Avg., as the task-specific signal overfits small calibration sets (Appendix~\ref{app:Ablation}).

\subsection{Differentiable Bit-Width Selection under Budget Constraints}

\begin{figure}[htb]   
  \centering
  \includegraphics[width=0.8\columnwidth]{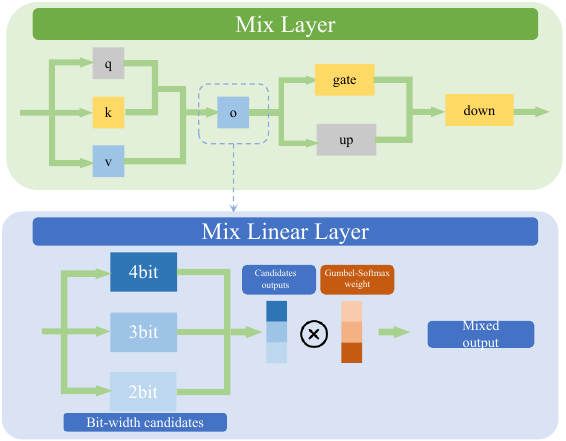} 
  \caption{\textbf{Mixed-precision layer and differentiable bit selection.}
\MethodName~ assigns a bit-width to each linear module by forming a weighted combination of pre-quantized candidates, where the weights are produced by a Gumbel--Softmax mask parameterized by trainable logits.}
  \label{fig:mix_layer}
\end{figure}

We optimize a continuous relaxation of the discrete assignment variables $z_{i,j,b}$ by learning, for each module $(i,j)$, a categorical distribution over candidate bit-widths $b \in \mathcal{B}$ (Figure~\ref{fig:mix_layer}).
Concretely, we parameterize probabilities $p_{i,j,b}\in[0,1]$ with $\sum_{b\in\mathcal{B}} p_{i,j,b}=1$
using logits $\pi_{i,j,b}$ and the Gumbel--Softmax reparameterization~\cite{jang2016categorical}:
\begin{equation}
\begin{aligned}
p_{i,j,b}
&=
\left[
\mathrm{softmax}\!\left(
\left\{\frac{\pi_{i,j,b'} + g_{i,j,b'}}{\tau}\right\}_{b'\in\mathcal{B}}
\right)
\right]_b, \\
g_{i,j,b}
&= -\log(-\log u_{i,j,b}), \qquad u_{i,j,b}\sim \mathcal{U}(0,1).
\end{aligned}
\end{equation}
where $\tau>0$ is the temperature and $\{g_{i,j,b}\}_{b\in\mathcal{B}}$ are i.i.d.\ Gumbel noises.

Replacing the binary variables $z_{i,j,b}$ by their continuous counterparts $p_{i,j,b}$ yields differentiable mixed-precision weights:
\begin{equation}
W^{(i,j)}_{\mathrm{mix}}
=
\sum_{b \in \mathcal{B}} p_{i,j,b}\, W^{(i,j)}_{b}.
\end{equation}

The expected average bit-width under $p$ is
\begin{equation}
\hat{b}(p)
=
\frac{
\sum_{i=1}^{L}\sum_{j \in \mathcal{J}} N_{i,j} \sum_{b \in \mathcal{B}} b \cdot p_{i,j,b}
}{
\sum_{i=1}^{L}\sum_{j \in \mathcal{J}} N_{i,j}
}.
\end{equation}

Given a target average bit-width $b_{\text{target}}$, we enforce $\hat{b}(p) = b_{\text{target}}$ via an augmented Lagrangian formulation, which---unlike the soft multiplicative penalty used in DNAS---dynamically tightens the constraint via learned dual variables, ensuring the expected bit-width converges precisely to the target before Stage~II discretization.
\begin{equation}
\mathcal{L}
=
\ell_{\mathrm{mse}}
+
\lambda _1\big(\hat{b}(p)-b_{\text{target}}\big)
+
\lambda _2\big(\hat{b}(p)-b_{\text{target}}\big)^2,
\label{eq:loss}
\end{equation}
where $\lambda _1, \lambda _2$ are two trainable parameters, and they will maximize the penalty term during the training process.
We update the mask parameters $\{\pi_{i,j,b}\}$ by minimizing $\mathcal{L}$ (using stochastic gradient descent on the calibration set).
This formulation yields stable end-to-end optimization of soft bit assignments under a \textbf{global expected bit-width budget}.
Replacing this augmented Lagrangian with a DNAS-style multiplicative penalty degrades performance by up to $-3.45$ Avg., because the soft penalty fails to converge the expected bit-width precisely to the target, forcing the subsequent integer program to make aggressive corrections (Appendix~\ref{app:Ablation}).
The final discrete model is obtained in Stage~II by projecting the learned probabilities to a budget-feasible binary assignment.

\subsection{Discrete Bit Assignments}

After optimization, the relaxed probabilities $p_{i,j,b}$ typically become highly peaked but remain continuous.
To obtain a deployable mixed-precision model, we convert these soft assignments into discrete bit-width selections while strictly respecting the global budget constraint.

\textbf{Assignment scores.}
For each module $(i,j)$ and candidate bit-width $b\in\mathcal{B}$, we define an assignment score
\begin{equation}
s_{i,j,b} = p_{i,j,b},
\end{equation}
which reflects the relative preference learned during constrained optimization.
Intuitively, higher scores correspond to bit-widths that better preserve hidden representations under the global budget pressure, as induced by the reconstruction objective and the augmented Lagrangian formulation.

\textbf{Discrete budget-constrained allocation.}
We select exactly one bit-width per module by solving the following integer optimization problem:
\begin{equation}
\label{eq:discrete_alloc}
\begin{aligned}
\max_{\{z_{i,j,b}\}} \quad &
\sum_{i,j}\sum_{b\in\mathcal{B}} z_{i,j,b}\, s_{i,j,b} \\
\text{s.t.} \quad &
\sum_{b\in\mathcal{B}} z_{i,j,b} = 1, \quad \forall (i,j), \\
&
\frac{
\sum_{i,j} N_{i,j} \sum_{b\in\mathcal{B}} b\, z_{i,j,b}
}{
\sum_{i,j} N_{i,j}
}
\le b_{\text{target}}, \\
&
z_{i,j,b} \in \{0,1\}.
\end{aligned}
\end{equation}

This formulation corresponds to a 0--1 knapsack-style problem with a single global resource constraint.
Although NP-hard in general, the problem size in our setting is moderate (one decision per linear module and bit-width) and can be solved efficiently using dynamic programming or standard integer linear programming solvers.
In practice, we obtain exact budget-feasible assignments with negligible overhead.
This discretization step decouples continuous importance estimation from discrete resource allocation, ensuring stable optimization and strict compliance.
Because the scores $s_{i,j,b}$ encode \textbf{module-level sensitivity} rather than budget-specific assignments, they can be reused: solving Eq.~\ref{eq:discrete_alloc} with a different $b_{\text{target}}$ yields a new allocation without retraining Stage~I.
Empirically, Pearson correlations between Stage~I soft scores and Stage~II discrete assignments consistently exceed 0.7 (often $>$0.9), confirming that the ILP largely preserves the learned preferences (Appendix~\ref{app:pearson}).

\section{Experiments}

\begin{table*}[ht]
\centering
\caption{\textbf{Mixed-precision quantization under target average bit budgets.}
Zero-shot accuracy (\%) on three LLMs at target average bit-widths of 2.5, 3.0, and 3.5.
\MethodName~ is compared with fixed-precision PTQ baselines AQLM and the search-based mixed-precision method AMQ.
All methods are evaluated under identical average bit budgets; ``Avg.'' denotes the average across tasks.
}
\label{tab:quantization_results}

\footnotesize

\renewcommand{\arraystretch}{0.92}

\resizebox{\linewidth}{!}
{
    \begin{tabular}{c|c|c||cccccc|c}
    \toprule
    \textbf{Model} & \makecell{\textbf{Avg.} \\ \textbf{Bits}} & \textbf{Method} & \textbf{ARC-e} & \textbf{ARC-c} & \textbf{PIQA} & \textbf{HellaS.} & \textbf{WinoG.} & \textbf{BoolQ} & \textbf{Avg.} \\
    \midrule
    
    \multirow{10}{*}{\makecell{Llama3.1-8B \\ -Instruct}} & 16 & FP16 & 92.24 & 83.39 & 78.62 & 75.85 & 61.80 & 85.75 & 79.60 \\ 
    \cline{2-10}
    
    & \multirow{3}{*}{2.5} 
    & AQLM  & 86.95 & 76.61 & 67.63 & 61.97 & 51.70 & 82.57 & 71.23 \\ 
    & 
    & AMQ & 87.65 & 77.63 & \textbf{72.63} & 66.33 & \textbf{56.04} & 81.90 & 73.69 \\ \rowcolor{gray!40}
    \cellcolor{white} & \cellcolor{white} 
    & GAMMA & \textbf{89.95} & \textbf{77.97} & 71.55 & \textbf{66.63} & 52.64 & \textbf{84.40} & \textbf{73.86} \\ 
    \cline{2-10}

    & \multirow{3}{*}{3.0} & 
    AQLM  & 85.71 & 71.19 & 67.90 & 60.33 & 52.49 & 81.90 & 69.92 \\ 
    & & 
    AMQ   & 89.77 & 81.02 & 77.69 & 69.88 & 58.88 & 82.87 & 76.68 \\ \rowcolor{gray!40}
    \cellcolor{white} & \cellcolor{white} & 
    GAMMA & \textbf{91.36} & \textbf{82.03} & \textbf{78.35} & \textbf{73.87} & \textbf{59.43} & \textbf{84.86} & \textbf{78.31} \\ 
    \cline{2-10}

    & \multirow{3}{*}{3.5} & 
    AQLM  & \textbf{92.59} & 84.07 & 71.93 & 65.89 & 58.09 & 84.74 & 76.21 \\ 
    & & 
    AMQ   & 91.71 & \textbf{85.42} & 76.88 & \textbf{74.74} & 58.72 & 84.86 & 78.72 \\ \rowcolor{gray!40}
    \cellcolor{white} & \cellcolor{white} & 
    GAMMA & 92.24 & 82.71 & \textbf{79.54} & 71.68 & \textbf{61.01} & \textbf{85.20} & \textbf{78.73} \\ 
    \cline{1-10}

    \multirow{10}{*}{Qwen3-8B} & 16 & FP16 & 88.54 & 84.07 & 79.54 & 84.65 & 68.98 & 86.06 & 81.97 \\ 
    \cline{2-10}
    
    & \multirow{3}{*}{2.5} & 
    AQLM  & 72.13 & 60.68 & 67.63 & 68.28 & 55.01 & 72.63 & 64.53 \\ 
    & & 
    AMQ   & 68.08 & 61.69 & 73.83 & \textbf{79.40} & 63.22 & \textbf{86.45} & 72.11 \\ \rowcolor{gray!40}
    \cellcolor{white} & \cellcolor{white} & 
    GAMMA & \textbf{83.95} & \textbf{72.88} & \textbf{75.73} & 77.72 & \textbf{64.48} & 85.87 & \textbf{76.77} \\ 
    \cline{2-10}

    & \multirow{3}{*}{3.0} & 
    AQLM  & 56.79 & 53.22 & 75.35 & 75.47 & 61.88 & 86.36 & 68.17 \\ 
    & & 
    AMQ   & 75.66 & 62.37 & 76.55 & \textbf{81.51} & 64.25 & 84.65 & 74.16 \\ \rowcolor{gray!40}
    \cellcolor{white} & \cellcolor{white} & 
    GAMMA & \textbf{91.18} & \textbf{80.00} & \textbf{80.90} & 80.95 & \textbf{68.43} & \textbf{85.54} & \textbf{81.16} \\ 
    \cline{2-10}

    & \multirow{3}{*}{3.5} & 
    AQLM  & 76.54 & 70.51 & 76.66 & 82.66 & 56.04 & 85.38 & 74.63 \\ 
    & & 
    AMQ   & 88.89 & 80.00 & 79.33 & 82.74 & 67.64 & \textbf{86.27} & 80.81 \\ \rowcolor{gray!40}
    \cellcolor{white} & \cellcolor{white} & 
    GAMMA & \textbf{89.07} & \textbf{82.03} & \textbf{79.60} & \textbf{83.62} & \textbf{68.43} & 86.06 & \textbf{81.46} \\ 
    \cline{1-10}

    \multirow{10}{*}{Qwen3-14B} & 16 & FP16 & 91.36 & 81.69 & 81.66 & 88.08 & 71.67 & 88.81 & 83.87 \\ 
    \cline{2-10}
    
    & \multirow{3}{*}{2.5} & 
    AQLM  & 86.77 & 82.37 & 75.84 & 77.51 & 66.14 & 85.14 & 78.96 \\ 
    & & 
    AMQ   & 89.59 & 77.97 & 79.33 & 86.35 & \textbf{69.46} & \textbf{88.32} & 81.83 \\ \rowcolor{gray!40}
    \cellcolor{white} & \cellcolor{white} & 
    GAMMA & \textbf{89.95} & \textbf{84.41} & \textbf{80.41} & \textbf{86.43} & 69.14 & \textbf{88.32} & \textbf{83.11} \\ 
    \cline{2-10}

    & \multirow{3}{*}{3.0} & 
    AQLM  & 83.95 & 77.63 & 79.43 & 85.05 & 65.51 & 87.49 & 79.84 \\ 
    & & 
    AMQ   & 89.59 & \textbf{81.69} & 80.69 & 86.59 & 70.64 & 88.81 & 83.00 \\ \rowcolor{gray!40}
    \cellcolor{white} & \cellcolor{white} & 
    GAMMA & \textbf{90.30} & \textbf{81.69} & \textbf{82.37} & \textbf{87.73} & \textbf{71.59} & \textbf{88.87} & \textbf{83.75} \\ 
    \cline{2-10}

    & \multirow{3}{*}{3.5} & 
    AQLM  & 90.30 & \textbf{83.05} & \textbf{82.97} & 85.19 & \textbf{72.06} & 88.44 & 83.66 \\ 
    & & 
    AMQ   & \textbf{91.01} & 80.68 & 81.56 & 87.22 & 71.74 & 88.23 & 83.40 \\ \rowcolor{gray!40}
    \cellcolor{white}  & \cellcolor{white}& 
    GAMMA & 90.48 & 82.03 & 81.88 & \textbf{87.98} & \textbf{71.74} & \textbf{88.90} & \textbf{83.84} \\
    
    \bottomrule
    \end{tabular}
}
\end{table*}

\subsection{Experimental Setup}
\label{sec:exp_setup}

\textbf{Models and Benchmarks.}
We evaluate GAMMA on Llama3.1-8B-Instruct~\cite{grattafiori2024llama}, Qwen3-8B~\cite{bai2023qwen}, and Qwen3-14B using zero-shot accuracy via OpenCompass, with additional scalability experiments on Qwen3-32B (Appendix~\ref{app:scalability_32b}).
The evaluation covers six standard benchmarks: ARC-e, ARC-c~\cite{clark2018think}, PIQA~\cite{bisk2020piqa}, HellaSwag~\cite{zellers2019hellaswag}, WinoGrande~\cite{sakaguchi2021winogrande}, and BoolQ~\cite{clark2019boolq}.

\textbf{Implementation Details.}
We focus on mixed-precision weight-only quantization with candidate bit-widths $\mathcal{B}=\{2,3,4\}$, utilizing pre-quantized candidates produced by QTIP. 
The differentiable masks are optimized on a calibration set $\mathcal{D}_{cal}$ comprising 1024 sequences (length 2048) sampled from RedPajama.
Detailed optimization hyper-parameters (e.g., learning rate, batch size) and decoding configurations are deferred to Appendix~\ref{app:Implementation_Details}.

\textbf{Baselines.}
We compare GAMMA against three representative methods: 
(i) \textbf{AQLM~\cite{egiazarian2024extreme}}, serving as a strong fixed-precision baseline; 
(ii) \textbf{AMQ}, a search-based layer-wise mixed-precision allocation method; 
and (iii) \textbf{QTIP} to isolate the benefits of mixed-precision allocation under the identical PTQ backbone.

\subsection{Main Results}
\label{sec:main_results}

We organize results around three questions: (1) does learned allocation outperform fixed-precision and search-based alternatives? (2) does the allocation scale smoothly across budgets? and (3) can mixed precision shift the operating point to lower memory footprints?
Additionally, generative benchmarks (GSM8K, RULER) in Appendix~\ref{app:Generalizability} confirm that GAMMA maintains reasoning performance beyond zero-shot classification.

\textbf{Learned allocation outperforms both fixed-precision and search-based alternatives.}
The advantage of \MethodName~ is most pronounced under extremely low average-bit budgets (2.5--3.0 bits), where quantization noise has a disproportionate effect and performance depends critically on where limited bits are allocated (Table~\ref{tab:quantization_results}).
Fixed-precision quantization (AQLM) cannot adapt to heterogeneous module sensitivity: on Qwen3-8B at 3.0 bits, AQLM reaches only 68.17 Avg.\ while \MethodName~ attains \textbf{81.16 (+12.99)}.
\MethodName~ also substantially outperforms the search-based baseline AMQ---by \textbf{+7.00} at 3.0 bits and \textbf{+4.66} at 2.5 bits on Qwen3-8B---indicating that the gains arise from more effective allocation, not mixed precision alone.
Importantly, the improvements are broad-based rather than driven by a single benchmark.
On Llama3.1-8B-Instruct at 3.0 bits, \MethodName~ outperforms AMQ on all six benchmarks (78.31 vs.\ 76.68 Avg.), with gains distributed across reasoning (ARC-c: 82.03 vs.\ 81.02), commonsense (HellaSwag: 73.87 vs.\ 69.88), and reading comprehension (BoolQ: 84.86 vs.\ 82.87).
On the larger Qwen3-14B, \MethodName~ maintains strong consistency, and further scales to Qwen3-32B with GAMMA-3.0bit (\textbf{85.97}) surpassing even FP16 (84.83; Appendix~\ref{app:scalability_32b}).

\begin{figure}[htbp]
    \centering
     \begin{subfigure}[b]{0.48\linewidth}
         \centering
         \includegraphics[width=\textwidth]{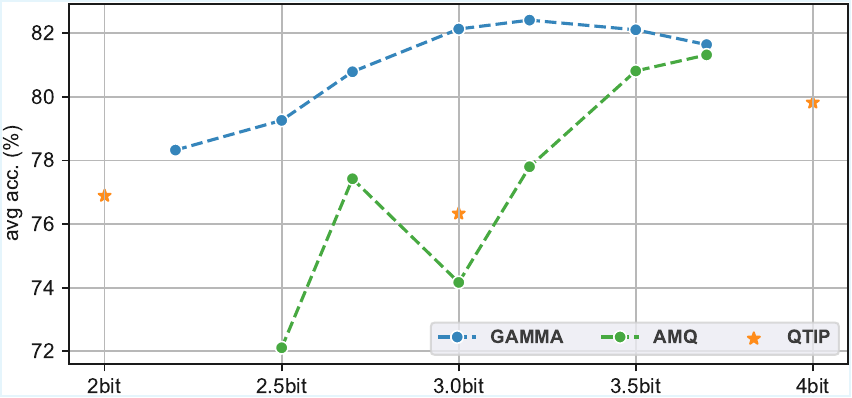}
         \caption{Qwen3-8B}
         \label{fig:qwen3-8b}
     \end{subfigure}
     \hfill
     \begin{subfigure}[b]{0.48\linewidth}
         \centering
         \includegraphics[width=\textwidth]{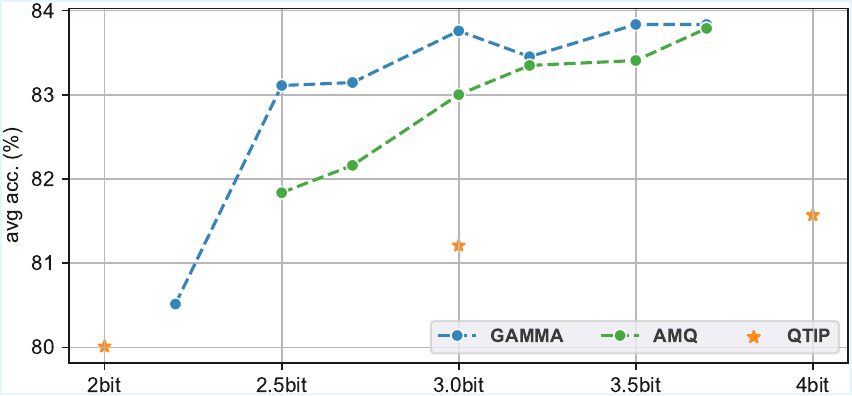}
         \caption{Qwen3-14B}
         \label{fig:qwen3-14b}
     \end{subfigure}
     
\caption{\textbf{Budget--accuracy scaling.}
Average zero-shot accuracy versus target average bit-width on Qwen3-8B and Qwen3-14B.}
    \label{fig:budget_curve}
\end{figure}


\textbf{The allocation scales smoothly across budgets.}
Figure~\ref{fig:budget_curve} visualizes the budget--accuracy trade-off.
\MethodName~ improves \textbf{smoothly and monotonically} as $b_{\text{target}}$ increases, indicating a stable allocation path rather than brittle, budget-specific solutions.
The curve shows substantial gains from 2.5 to 3.0 bits, followed by smaller improvements toward 3.5 bits, suggesting diminishing returns once the most sensitive modules have been promoted.
Across the 2--4 bit range, \MethodName~ \textbf{outperforms AMQ at every matched budget}, and can even slightly surpass the fixed 4-bit QTIP baseline---indicating that \textbf{better bit placement can matter more than uniformly increasing precision}.

\begin{table}[b]
\centering
\caption{Fixed-budget comparison on the same QTIP PTQ backbone. \MethodName~ outperforms fixed 3-bit QTIP at $b_{\text{target}}{=}3.0$ and remains competitive at $b_{\text{target}}{=}2.5$, demonstrating improved memory efficiency. ``Avg.'' is the average across tasks.}
\label{tab:fix_size_results}

\renewcommand{\arraystretch}{0.92}

\resizebox{\linewidth}{!}
{
    \begin{tabular}{c|c|c||cccccc|c}
    \toprule
    \textbf{Model} & \makecell{\textbf{Avg.} \\ \textbf{Bits}} & \textbf{Method} & \textbf{Arc-e} & \textbf{Arc-c} & \textbf{PIQA} & \textbf{HellaS.} & \textbf{WinoG.} & \textbf{BoolQ}  & \textbf{Avg.} \\
    \midrule

    \multirow{2}{*}{\makecell{Llama3.1-8B \\ -Instruct}} 
    & \multirow{2}{*}{3.0} 
    & QTIP & 91.18 & 84.07 & 77.80 & 69.46 & 57.93 & 85.02 & 77.58 \\ 
    & 
    & GAMMA & 91.36 & 82.03 & 78.35 & 73.87 & 59.43 & 84.86 & 78.31  \\
    
    \cline{1-10}

    \multirow{2}{*}{\makecell{Qwen3-8B}} 
    & \multirow{2}{*}{3.0} 
    &  QTIP & 85.01 & 74.58 & 78.62 & 67.76 & 66.77 & 85.26 & 76.33 \\ 
    & 
    & GAMMA & 91.18 & 80.00 & 80.90 & 80.95 & 68.43 & 85.54 & 81.16 \\
        \cline{2-10}
    & 2.5 & GAMMA & \textbf{83.95} & \textbf{72.88} & \textbf{75.73} & 77.72 & \textbf{64.48} & 85.87 & \textbf{76.77} \\

    \cline{1-10}

    \multirow{3}{*}{\makecell{Qwen3-14B}} 
    & \multirow{2}{*}{3.0} 
    &  QTIP & 88.18 & 81.02 & 81.72 & 78.05 & 69.85 & 88.41 & 81.21  \\ 
    & 
    &GAMMA & 90.30 & 81.69 & 82.37 & 87.73 & 71.59 & 88.87  & 83.75 \\
    \cline{2-10}
    & 2.5 & GAMMA & \textbf{89.95} & \textbf{84.41} & \textbf{80.41} & \textbf{86.43} & 69.14 & \textbf{88.32} & \textbf{83.11} \\
    
    \bottomrule
    \end{tabular}
}
\end{table}


\textbf{Mixed precision shifts the operating point to lower memory footprints.}
Table~\ref{tab:fix_size_results} compares \MethodName~ against a fixed-precision QTIP baseline under the same PTQ backbone, isolating the benefit of mixed-precision allocation.
At $b_{\text{target}}{=}3.0$, \MethodName~ consistently outperforms fixed 3-bit QTIP on all three models, confirming that the gains come from heterogeneous allocation, not from changing the quantizer.
More importantly, \MethodName~ enables comparable capability at substantially lower average precision.
On Qwen3-8B, \MethodName~ at \textbf{2.5 bits reaches 76.77}, essentially matching fixed 3-bit QTIP (76.33) with fewer bits.
On Qwen3-14B, \MethodName~ at \textbf{2.5 bits attains 83.11}, surpassing the fixed 3-bit baseline (81.21).
These results demonstrate that mixed-precision allocation can \textbf{shift the operating point}, achieving similar performance at a smaller memory footprint.

\section{Analysis}

\begin{figure}[htbp]
    \centering
     \begin{subfigure}[b]{0.48\linewidth}
         \centering
         \includegraphics[width=\linewidth]{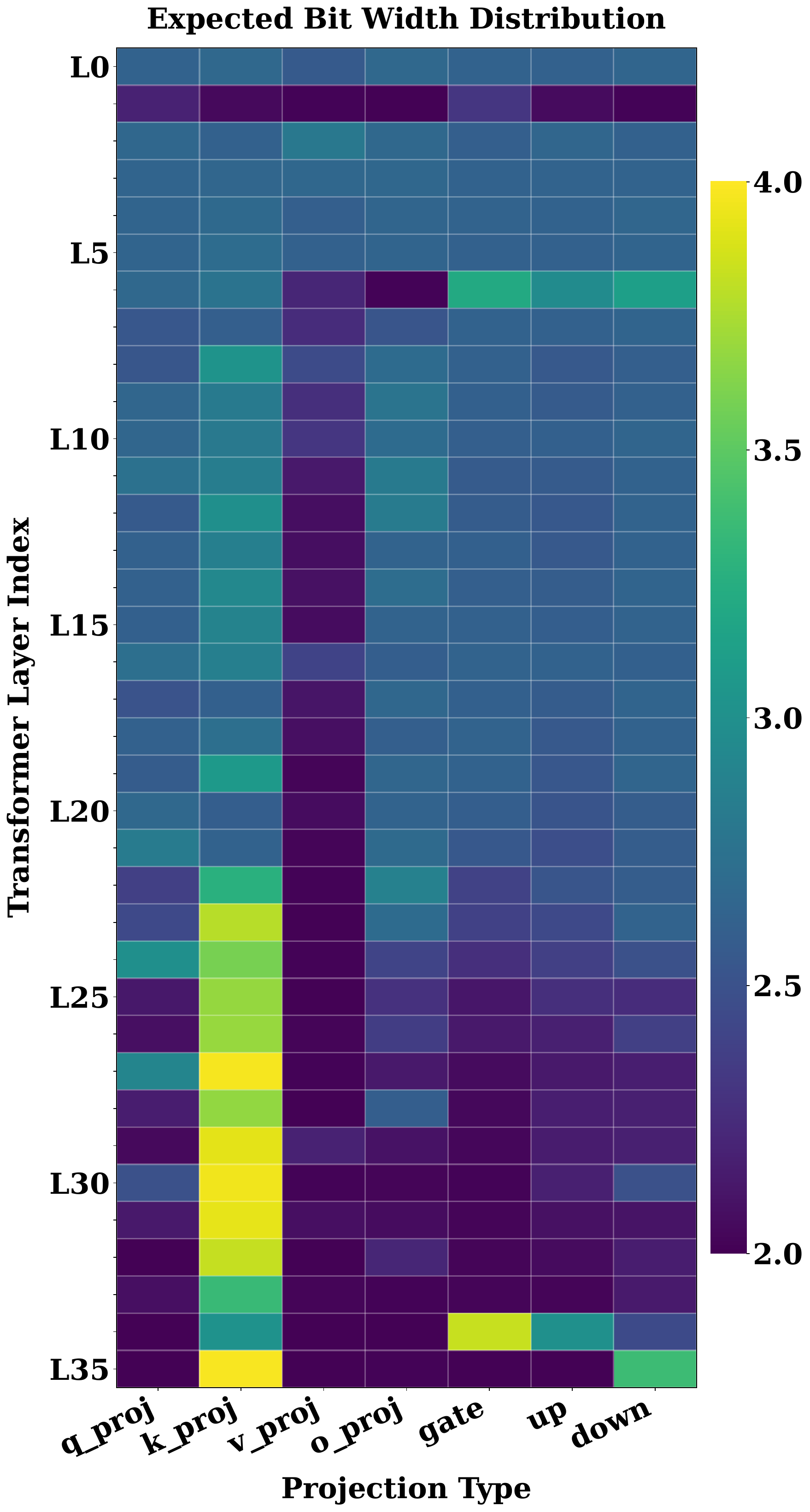}
         \caption{RedPajama $b_{\text{target}}{=}2.5$}
         \label{fig:red25}
     \end{subfigure}
     \hfill
     \begin{subfigure}[b]{0.48\linewidth}
         \centering
         \includegraphics[width=\linewidth]{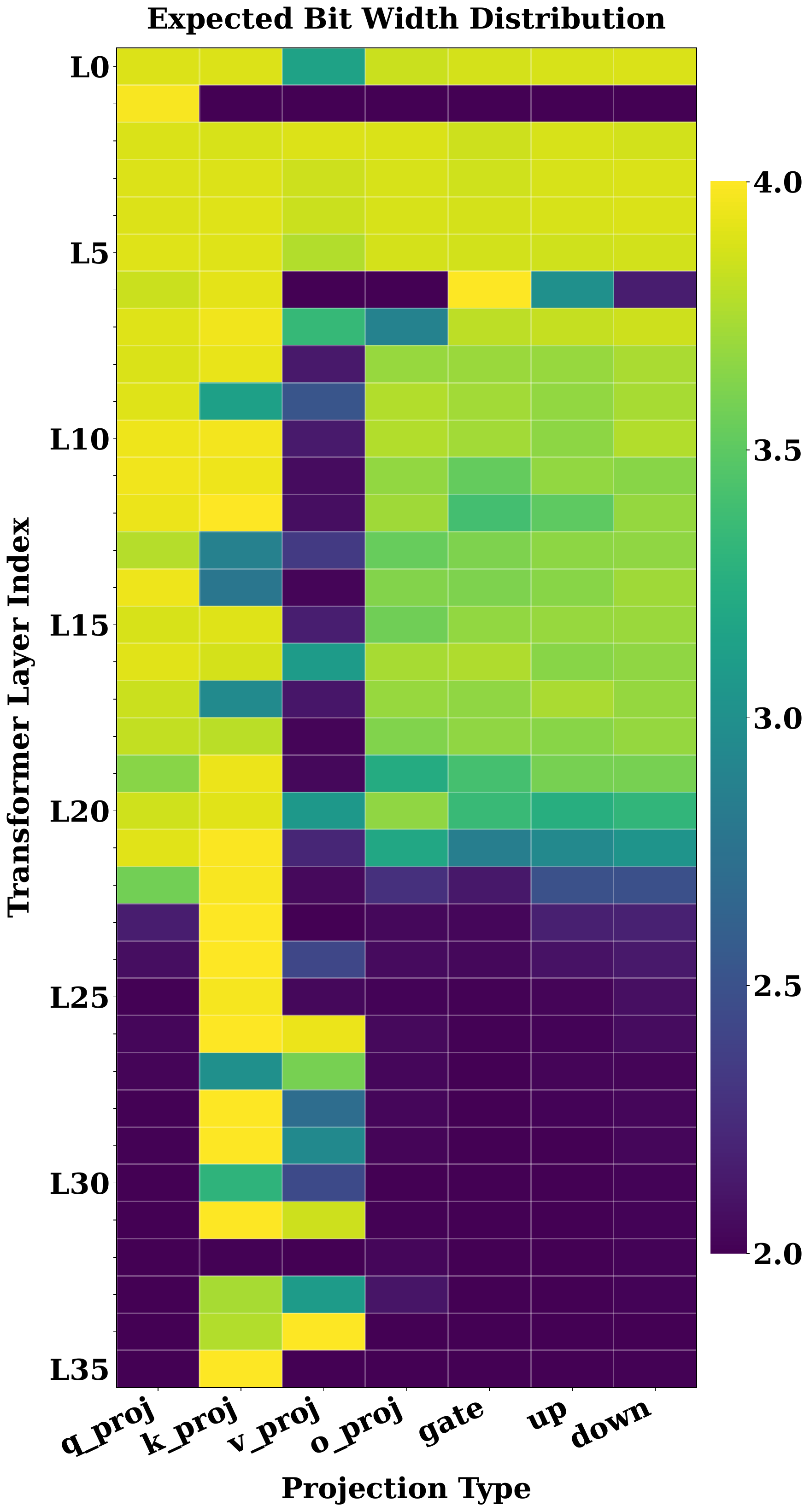}
         \caption{RedPajama $b_{\text{target}}{=}3.0$}
         \label{fig:red30}
     \end{subfigure}
\caption{
\textbf{Learned bit-width allocation patterns across budgets.}
Heatmaps show the expected bit-width assigned by \MethodName~ to each projection type (x-axis) in every Transformer layer (y-axis), optimized on RedPajama under target budgets $b_{\text{target}}\in\{2.5,3.0\}$.
The two allocations are highly consistent (cosine similarity 0.9755), indicating that the learned pattern largely preserves the same relative sensitivity structure across budgets.
}
\label{fig:heatmaps}
\end{figure}

\subsection{Learned bitwidth patterns across budgets and calibration sets}
\label{sec:analysis_pattern}

Figure~\ref{fig:heatmaps} reveals a structured, non-uniform allocation that is \textbf{largely preserved} when the target budget changes (cosine similarity \textbf{0.9755} between $b_{\text{target}}{=}2.5$ and $3.0$).
Rather than producing qualitatively different assignments at each budget, \MethodName~ expands or contracts a common pattern: budget increases primarily upgrade additional modules along the same sensitivity ordering, rather than switching priorities.

Within this pattern, k\_proj consistently receives higher precision than other projections---plausibly because errors in the key pathway are amplified through attention scoring---while v\_proj is kept near the minimum precision.

The same structure holds across calibration distributions: RedPajama and WikiText allocations at $b_{\text{target}}{=}3.0$ have cosine similarity \textbf{0.9913} (Appendix~\ref{app:heatmaps}).
This stability is the foundation of score reuse: because the learned scores reflect an \textbf{intrinsic, dataset-invariant sensitivity ranking}, they can be projected to new budgets without retraining.



\subsection{Generalization of \MethodName}
\label{sec:analysis_generalization}

\begin{table}[ht]
\centering
\caption{\textbf{Generalization across Candidate Sets and Target Bit-Width Allocation}
Average zero-shot accuracy over the six core benchmarks.
Ternary/binary denote candidate sets $\{2,3,4\}$ and $\{2,4\}$, respectively.
Score-reuse reuses scores learned at $b_{\text{target}}{=}3.0$ and projects them to other budgets without retraining.}
\label{tab:generalization}

\renewcommand{\arraystretch}{0.92}

\resizebox{\linewidth}{!}
{
    \begin{tabular}{c|ccccc}

    \toprule
    & 2.5bit & 2.7bit & 3.0bit & 3.2bit & 3.5bit \\
    \midrule
    \MethodName~ (ternary) & 79.36 & 80.78 & 82.28 & 82.41 & 82.09 \\
    \MethodName~ (binary) & 77.52 & 80.66 & 81.90 & 80.87 & 82.07 \\
    \MethodName~ (score-reuse) & 79.18 & 79.77 & - & 80.96 & 81.36 \\
    AMQ (ternary) & 72.11 & 77.42 & 74.16 & 77.80 & 80.81 \\
    \bottomrule
    
    \end{tabular}
}
\end{table}

Table~\ref{tab:generalization} studies two complementary forms of generalization for \MethodName.
First, \MethodName~ is flexible to the candidate precision set.
For the binary candidate set $\{2,4\}$, we use a sigmoid relaxation (a two-class special case) instead of Gumbel--Softmax.
The ternary candidate set $\{2,3,4\}$ is consistently better than the binary set $\{2,4\}$ at the same budgets.
This indicates that, under tight constraints, distributing bits across more modules via moderate upgrades (2$\rightarrow$3) can be more effective than concentrating the budget on a few large upgrades (2$\rightarrow$4), since it reduces widespread low-precision bottlenecks.

Second, \MethodName~ exhibits \textbf{cross-budget score reuse} through \MethodName~(score-reuse).
Reusing the preference scores learned at $b_{\text{target}}{=}3.0$ yields competitive performance at other target budgets without retraining (Table~\ref{tab:generalization}).
Although score-reuse is slightly below budget-aligned training, it remains strong and \textbf{consistently outperforms AMQ} across all evaluated budgets.
This confirms that the learned scores encode a stable sensitivity ranking.
As a result, a single training run can serve multiple target budgets through the projection step (\textbf{a few minutes per budget} vs.\ hours for retraining or search), without re-optimizing masks for every deployment point.

\textbf{\MethodName~ is more time- and compute-efficient in practice.}
\MethodName~ offers superior time and compute efficiency during both quantization and inference. 
On an NVIDIA A100 GPU for Qwen3-8B, GAMMA requires approximately 1 hour to learn preference scores---a duration comparable to the Hessian trace computation in HAWQ-V2 but significantly faster than the 4-hour search pipeline required by AMQ.
Despite similar computational cost, GAMMA substantially outperforms a HAWQ-V2 re-implementation by +1.36 to +5.92 Avg.\ on Qwen3-8B across all budgets, confirming that learned scores capture richer sensitivity structure than static Hessian metrics (detailed comparison in Appendix~\ref{app:Quantizationtime}). 
Furthermore, unlike search-based methods that require re-computation for every new budget, GAMMA's learned sensitivity scores can be projected to arbitrary targets with negligible cost. 
At deployment, GAMMA’s inter-layer allocation allows each module to utilize pre-optimized kernels directly, maintaining a decoding throughput within $3\%$ of uniformly quantized baselines with no meaningful inference overhead, as shown in Appendix~\ref{app:Inference}.

\textbf{Each design choice contributes independently.}
We ablate the two key components of \MethodName~ on Qwen3-8B to quantify their individual contributions (full results in Appendix~\ref{app:Ablation}).
Replacing the augmented Lagrangian with a DNAS-style multiplicative penalty---which scales the loss by a soft function of the budget deviation rather than enforcing exact convergence---degrades performance by \textbf{$-$2.56 to $-$3.45} Avg.: without precise budget convergence, the subsequent ILP is forced into aggressive corrections that degrade allocation quality.
Replacing the hidden-state reconstruction loss with a standard cross-entropy loss degrades performance by \textbf{$-$0.95 to $-$3.29} Avg.: the end-to-end task loss conflates errors from all layers, providing a weaker gradient signal for per-module bit-width selection, and is more prone to overfitting on limited calibration data.
Both components are necessary; removing either one leads to consistent degradation across all evaluated budgets.

\section{Conclusion}

We introduced \MethodName, a quantizer-agnostic framework for global mixed-precision bit allocation under arbitrary average-bit budgets.
\MethodName~ learns \textbf{module-wise precision preferences without weight retraining}, produces \textbf{exact budget-feasible} discrete assignments via integer programming, and yields \textbf{reusable allocation scores} that transfer across deployment targets.
Across Llama and Qwen models in the extremely low-bit regime, \MethodName~ improves the budget--accuracy trade-off over strong baselines, and can match fixed 3-bit QTIP quality at lower average-bit budgets.
Analysis further shows that the learned allocation pattern is stable across budgets and calibration sets, enabling \textbf{score reuse} that reduces per-budget adaptation from hours to a few minutes.

\section*{Limitations}
\label{app:limitation}

\MethodName~assumes access to pre-quantized candidate weights at multiple bit-widths; its applicability and quality therefore depend on the availability and strength of the underlying PTQ backbone.
While the projection step is typically inexpensive at our scale, solving the discrete assignment may become more costly as the candidate set or module granularity grows.
Our evaluation covers zero-shot classification, mathematical reasoning (GSM8K), and long-context tasks (RULER); extending to broader generative benchmarks such as code generation and open-ended dialogue remains future work.

\bibliography{custom}

\appendix

\clearpage

\section{Generalizability to downstream tasks}
\label{app:Generalizability}

\begin{table}[ht]
\centering
\caption{Model performance on downstream tasks across various bit-width configurations. Even on complex tasks, the performance scales smoothly and monotonically with the bit budget, demonstrating robust generalizability across diverse tasks.}
\label{tab:general_tasks}

\renewcommand{\arraystretch}{0.92}

\resizebox{\linewidth}{!}
{
    \begin{tabular}{c|c||cc||cc}
    \toprule
    \textbf{Model} & \makecell{\textbf{Config}} & \textbf{GSM8K(acc.)↑} & \textbf{RULER(acc.)↑} & \textbf{Wiki2(ppl.)↓} & \textbf{C4(ppl.)↓} \\
    \midrule

    \multirow{4}{*}{\makecell{Llama3.1-8B \\ -Instruct}} 
    & 16bit & 84.26 & 86.12 & 7.22 & 11.39 \\ 
        \cline{2-6}
    & GAMMA-3.5bit & 81.73 & 85.62 & 8.04 & 12.76 \\
    & GAMMA-3.0bit & 80.14 & 84.48 & 8.11 & 12.87 \\
    & GAMMA-2.5bit & 61.03 & 79.21 & 9.12 & 14.50 \\
    
    \cline{1-6}

    \multirow{4}{*}{\makecell{Qwen3-8B}} 
    & 16bit & 93.42 & 87.95 & 9.72 & 15.43 \\ 
        \cline{2-6}
    & GAMMA-3.5bit & 92.87 & 88.52 & 10.71 & 16.56 \\
    & GAMMA-3.0bit & 89.61 & 85.72 & 11.03 & 16.97 \\
    & GAMMA-2.5bit & 88.25 & 85.07 & 11.20 & 17.03 \\

    \cline{1-6}

    \multirow{4}{*}{\makecell{Qwen3-14B}} 
    & 16bit & 94.98 & 92.75 & 8.65 & 13.82  \\ 
        \cline{2-6}
    & GAMMA-3.5bit & 94.47 & 92.32 & 9.26 & 14.55 \\
    & GAMMA-3.0bit & 93.10 & 90.56 & 9.52 & 14.90 \\
    & GAMMA-2.5bit & 90.83 & 89.43 & 9.76 & 15.22 \\
    
    \bottomrule
    \end{tabular}
}
\end{table}

Table~\ref{tab:general_tasks} presents the performance of \MethodName~ on complex downstream tasks across various bit-width configurations.
Although the bit-width allocation is calibrated exclusively on the predominantly text-based RedPajama dataset, minimizing the layer-wise reconstruction error effectively preserves inter-layer semantic integrity. 
Consequently, the model maintains strong generalization capabilities across diverse downstream tasks. Extensive evaluations on GSM8K (mathematical reasoning), RULER (long-context modeling up to 32K tokens), and standard perplexity benchmarks (WikiText2 and C4) validate this robustness. 
Notably, operating at an ultra-low precision of 3.0–3.5 bits, the quantized Qwen models retain $>95\%$ of their full-precision (FP16) performance on GSM8K and RULER, alongside a graceful degradation in perplexity.

\section{Inference throughput evaluation}
\label{app:Inference}

\begin{table}[htp]
\centering
\caption{Comparison of inference throughput between mixed-precision and uniform-precision quantized models. GAMMA utilizes the AQLM as its quantization backend and performs inference using AQLM's Triton kernels.}
\label{tab:infer_efficiency}

\renewcommand{\arraystretch}{0.92}

{
    \begin{tabular}{ccc}
    \toprule
    \textbf{Model} & \makecell{\textbf{Config}} & \textbf{tokens/s↑} \\
    \midrule

    \multirow{6}{*}{\makecell{Qwen3-8B}} 
    & AQLM-4bit & 18.95 \\ 
    & AQLM-3bit & 18.58 \\
    & AQLM-2bit & 18.45 \\
    & GAMMA-3.5bit & 18.95 \\ 
    & GAMMA-3.0bit & 18.11 \\
    & GAMMA-2.5bit & 18.49 \\
    
    \cline{1-3}

    \multirow{6}{*}{\makecell{Qwen3-14B}} 
    & AQLM-4bit & 16.67 \\ 
    & AQLM-3bit & 16.72 \\
    & AQLM-2bit & 16.88 \\
    & GAMMA-3.5bit & 16.78 \\ 
    & GAMMA-3.0bit & 16.45 \\
    & GAMMA-2.5bit & 16.68 \\
    
    \bottomrule
    \end{tabular}
}
\end{table}

Table~\ref{tab:infer_efficiency} presents a comparison of inference speeds between the GAMMA mixed-precision models, backed by the AQLM quantization method, and the uniformly quantized AQLM baselines. 
Throughput evaluations were performed using the standard AQLM benchmarking utility (generate\_benchmark.py) backed by Triton kernels on A100 GPUs. 
We measured the decoding speed using a batch size of 1 and an output generation length of 128, reporting the average over 10 consecutive trials.
By design, GAMMA enforces strictly inter-layer allocation, allowing each module to operate uniformly at a single bit-width and directly reuse pre-optimized VQ kernels without custom mixed-precision support. 
As a result, GAMMA achieves a decoding throughput within $3\%$ of the uniform AQLM models at any given bit-width, empirically validating that the proposed mixed-precision allocation introduces no meaningful inference overhead.

\section{Quantization-Time Efficiency Comparison}
\label{app:Quantizationtime}

\begin{table}[ht]
\centering
\caption{Comparison of quantization time between HAWQ-V2 and \MethodName~.}
\label{tab:quant_speed}

\renewcommand{\arraystretch}{0.92}

{
    \begin{tabular}{ccc}
    \toprule
    \textbf{Model} & \textbf{HAWQ-V2 (h)} & \textbf{GAMMA (h)} \\
    \midrule

    \textbf{Llama3.1-8B} & 1.28 & 1.33 \\
    \textbf{Qwen3-8B} & 1.63 & 1.51 \\
    \textbf{Qwen3-14B} & 2.88 & 2.43 \\
    
    \bottomrule
    \end{tabular}
}
\end{table}

\begin{table}[ht]
\centering
\caption{Performance comparison of quantized models between HAWQ-V2 and \MethodName~.}
\label{tab:quant_result}

\renewcommand{\arraystretch}{0.92}

{
    \begin{tabular}{cccc}
    \toprule
    \textbf{Config} & \textbf{3.5bit} & \textbf{3.0bit} & \textbf{2.5bit} \\
    \midrule

    \textbf{HAWQ-V2-QTIP} & 78.43 & 75.24 & 75.41 \\
    \textbf{GAMMA-QTIP} & \textbf{81.46} & \textbf{81.16} & \textbf{76.77} \\
    
    \bottomrule
    \end{tabular}
}
\end{table}

\begin{table*}[htbp]
\centering
\caption{Ablation study on the optimization objectives. ‘Simpler-Penalty’ denotes the replacement of the bit-width constraint term, while ‘CE-Loss’ indicates the substitution of the layer-wise teacher-forced reconstruction loss.}
\label{tab:Ablation}

\renewcommand{\arraystretch}{0.92}

{
    \begin{tabular}{c|c||ccccccc}
    \toprule
    \makecell{\textbf{Avg.} \\ \textbf{Bits}} & \makecell{\textbf{Config}} & \textbf{Arc-e} & \textbf{Arc-c} & \textbf{PIQA} & \textbf{HellaS.} & \textbf{WinoG.} & \textbf{BoolQ}  & \textbf{Avg.} \\
    \midrule

    \multirow{3}{*}{\makecell{3.5bit}} 
    & GAMMA & \textbf{89.07} & \textbf{82.03} & 79.60 & \textbf{83.62} & \textbf{68.43} & \textbf{86.06} & \textbf{81.46} \\
    & Simpler-Penalty & 87.48 & 78.64 & 79.43 & 74.42 & 62.98 & 85.14 & 78.01 \\
    & CE-Loss & 86.42 & 76.61 & \textbf{80.63} & 72.69 & 66.61 & 86.06 & 78.17 \\
    
    \cline{1-9}

    \multirow{3}{*}{\makecell{3.0bit}} 
    & GAMMA & \textbf{91.18} & 80.00 & \textbf{80.90} & \textbf{80.95} & \textbf{68.43} & 85.54 & \textbf{81.16} \\
    & Simpler-Penalty & 87.65 & 75.93 & 79.11 & 75.17 & 66.77 & \textbf{86.97} & 78.60 \\
    & CE-Loss & 89.36 & \textbf{83.39} & 78.35 & 76.48 & 67.01 & 86.70 & 80.21 \\

    \cline{1-9}

    \multirow{3}{*}{\makecell{2.5bit}} 
    & GAMMA & 83.95 & 72.88 & 75.73 & \textbf{77.72} & \textbf{64.48} & \textbf{85.87} & \textbf{76.77} \\
    & Simpler-Penalty & \textbf{87.48} & \textbf{76.61} & \textbf{75.90} & 69.34 & 62.19 & 85.11 & 76.10 \\
    & CE-Loss & 86.42 & 74.92 & 75.41 & 66.90 & 62.51 & 84.95 & 75.18 \\
    
    \bottomrule
    \end{tabular}
}
\end{table*}

Hessian-based parameter sensitivity evaluation for bit-width allocation is conventionally regarded as highly efficient. 
To critically examine this assumption, we re-implemented HAWQ-V2 \cite{dong2020hawq} —a classic mixed-precision method from the computer vision domain—for LLMs, utilizing Hutchinson's algorithm for trace computation combined with Integer Linear Programming (ILP). 
Contrary to the prevailing belief, our empirical analysis reveals that computing Hessian traces for LLMs incurs a computational cost comparable to Stage I of \MethodName~ (see Table~\ref{tab:quant_speed}). 
In practice, \MethodName~ actually executes faster than HAWQ-V2 on both Qwen3-8B and Qwen3-14B models. 
Regarding memory footprint, both methods exhibit comparable overhead, requiring roughly 2–3x and ~3x the model weight size, respectively. 
Crucially, despite this similar resource consumption, \MethodName~ substantially outperforms HAWQ-V2, achieving an average improvement of +1.36 to +5.92 on Qwen3-8B (see Table~\ref{tab:quant_result}). 
Furthermore, a decisive advantage of GAMMA is that its learned sensitivity scores are budget-agnostic and can be reused across various target bit-widths (Table~\ref{tab:generalization}). 
This effectively amortizes the one-time training cost, whereas HAWQ-V2 necessitates a complete re-computation of traces for each specific target budget.

\section{Ablation study on the optimization targets}
\label{app:Ablation}

To investigate the necessity of both the Augmented Lagrangian term and the reconstruction loss in our bit-width allocation objective (Eq.~\ref{eq:loss}), we conduct targeted ablation studies. Specifically, we evaluate our method against two variants:
\begin{itemize}
\item[$\bullet$] Simpler-Penalty: We replace the AL term with a DNAS-style multiplicative penalty, formulated as $L = \ell _{\text{mse}} \times \beta(\log((\hat{b}-b_{\text{target}})^2))^\gamma$, while keeping the original reconstruction loss unchanged.
\item[$\bullet$] CE-Loss: We substitute the layer-wise teacher-forced reconstruction loss with a standard end-to-end Cross-Entropy (CE) loss applied to the final output, while maintaining the AL constraint.
\end{itemize}
The ablation results in Table~\ref{tab:Ablation} demonstrate that \MethodName~ outperforms the Simpler-Penalty variant by +2.56 to +3.45 on average. 
We attribute this to the DNAS-style penalty's inability to precisely converge to the target budget during continuous search. 
Such misalignments force the subsequent ILP step into aggressive, sub-optimal projections, causing detrimental distribution shifts. 
Conversely, our Augmented Lagrangian dynamically enforces strict budget convergence via learned dual variables before ILP, eliminating projection mismatch.

Additionally, the CE-Loss variant degrades by -0.95 to -3.29 on average, verifying that our reconstruction objective independently contributes to performance by aligning hidden states with the full-precision teacher, avoiding overfitting to task-specific signals on small calibration data. 
Together, these findings confirm that both components---exact budget enforcement via AL and layer-wise reconstruction for stable sensitivity estimation---provide independent, complementary contributions to \MethodName's final performance.

\section{Quantizer-Agnostic Validation: AQLM Backend}
\label{app:aqlm_backend}

To validate that \MethodName's allocation quality is not tied to QTIP-specific properties, we conduct experiments using AQLM as an alternative PTQ backend.

\begin{table}[ht]
\centering
\caption{Zero-shot accuracy with AQLM as the PTQ backend. \MethodName~ consistently improves over uniform AQLM across all models and budgets.}
\label{tab:aqlm_backend}

\renewcommand{\arraystretch}{0.92}

\resizebox{\linewidth}{!}
{
    \begin{tabular}{c|c||cccccc|c}
    \toprule
    \textbf{Model} & \makecell{\textbf{Config}} & \textbf{ARC-e} & \textbf{ARC-c} & \textbf{PIQA} & \textbf{HellaS.} & \textbf{WinoG.} & \textbf{BoolQ} & \textbf{Avg.} \\
    \midrule

    \multirow{4}{*}{\makecell{Llama3.1-8B \\ -Instruct}} 
    & AQLM-3bit & 90.83 & 84.75 & 74.70 & 66.39 & 58.09 & 85.57 & 76.72 \\
    & GAMMA-3.5bit & 93.47 & 82.03 & 77.37 & 73.62 & 62.90 & 85.96 & 79.22 \\
    & GAMMA-3.0bit & 91.01 & 82.71 & 76.33 & 70.61 & 59.43 & 84.46 & 77.42 \\
    & GAMMA-2.5bit & 85.89 & 74.58 & 72.96 & 67.12 & 55.09 & 82.97 & 73.10 \\
    
    \cline{1-9}

    \multirow{3}{*}{\makecell{Qwen3-8B}} 
    & AQLM-3bit & 88.89 & 83.39 & 78.35 & 68.22 & 58.25 & 85.17 & 77.04 \\
    & GAMMA-3.5bit & 85.19 & 73.90 & 78.24 & 73.14 & 69.77 & 85.50 & 77.62 \\
    & GAMMA-3.0bit & 89.77 & 78.31 & 75.19 & 75.45 & 67.64 & 84.37 & 78.45 \\
    
    \cline{1-9}

    \multirow{3}{*}{\makecell{Qwen3-14B}} 
    & AQLM-3bit & 85.01 & 79.32 & 80.03 & 68.05 & 58.72 & 88.65 & 76.63 \\
    & GAMMA-3.5bit & 90.48 & 82.03 & 81.99 & 82.19 & 70.80 & 88.81 & 82.71 \\
    & GAMMA-3.0bit & 91.01 & 82.71 & 85.31 & 82.79 & 73.95 & 87.80 & 83.92 \\
    
    \bottomrule
    \end{tabular}
}
\end{table}

\MethodName~ consistently improves over uniform AQLM across all models and budgets (+1.4 to +7.3 Avg.), confirming that the allocation quality is backend-agnostic (Table~\ref{tab:aqlm_backend}).

\begin{table}[htbp]
\centering
\caption{Pearson correlation between Stage~I soft scores and Stage~II discrete ILP assignments.}
\label{tab:pearson}

\renewcommand{\arraystretch}{0.92}

{
    \begin{tabular}{c|c}
    \toprule
    \textbf{Config} & \textbf{Pearson $r$} \\
    \midrule

    Llama3-8B 2.5bit & 0.746 \\
    Llama3-8B 3.0bit & 0.781 \\
    Llama3-8B 3.5bit & 0.832 \\
    Qwen3-8B 3.0bit & 0.807 \\
    Qwen3-8B 3.5bit & 0.953 \\
    Qwen3-14B 2.5bit & 0.819 \\
    Qwen3-14B 3.0bit & 0.943 \\
    Qwen3-14B 3.5bit & 0.988 \\
    
    \bottomrule
    \end{tabular}
}
\end{table}

\section{Scalability: Qwen3-32B}
\label{app:scalability_32b}

\begin{table*}[htbp]
\centering
\caption{Zero-shot accuracy on Qwen3-32B. \MethodName-3.0bit surpasses both FP16 and uniform QTIP-3bit.}
\label{tab:qwen32b}

\renewcommand{\arraystretch}{0.92}

{
    \begin{tabular}{c||cccccc|c}
    \toprule
    \makecell{\textbf{Config}} & \textbf{ARC-e} & \textbf{ARC-c} & \textbf{PIQA} & \textbf{HellaS.} & \textbf{WinoG.} & \textbf{BoolQ} & \textbf{Avg.} \\
    \midrule

    16bit & 90.30 & 85.76 & 82.75 & 86.39 & 75.93 & 87.86 & 84.83 \\
    QTIP-3bit & 93.12 & 86.10 & 83.35 & 86.57 & 74.11 & 87.55 & 85.13 \\
    GAMMA-3.5bit & 93.47 & 89.83 & 83.30 & 87.69 & 76.09 & 87.55 & 86.32 \\
    GAMMA-3.0bit & 93.83 & 89.83 & 84.39 & 86.38 & 73.95 & 87.49 & 85.97 \\
    GAMMA-2.5bit & 91.53 & 85.08 & 83.57 & 85.82 & 73.32 & 88.47 & 84.63 \\
    
    \bottomrule
    \end{tabular}
}
\end{table*}

\MethodName~ scales well to 32B parameters: GAMMA-3.0bit (85.97) surpasses both FP16 (84.83) and uniform QTIP-3bit (85.13), and even GAMMA-2.5bit (84.63) nearly matches FP16 (Table~\ref{tab:qwen32b}).

\section{Stage I--II Alignment: Pearson Correlation}
\label{app:pearson}

To quantify how much Stage~II corrects Stage~I, we compute the Pearson correlation between Stage~II's discrete ILP assignments and Stage~I's learned Gumbel-Softmax scores (Table~\ref{tab:pearson}).

Correlations consistently exceed 0.7 (often $>$0.9), confirming that Stage~II predominantly selects the bit-width with the highest learned score. The ILP serves as a principled projection ensuring exact budget compliance while largely preserving the learned preferences.

\section{Extended Implementation Details}
\label{app:Implementation_Details}

This section details the optimization hyperparameters, calibration data configuration, and evaluation protocols utilized within the GAMMA framework.

\textbf{Optimization Parameters (Stage I)}
\begin{itemize}
    \item The categorical mask parameters are optimized using stochastic gradient descent on a held-out calibration set $\mathcal{D}_{\mathrm{cal}}$.
    \item $\mathcal{D}_{\mathrm{cal}}$ contains 1024 sequences sampled from RedPajama~\cite{weber2024redpajama} with sequence length 2048.
    \item Stage~I uses a batch size of 8 and runs for 1120 optimization steps with learning rate $5\times10^{-3}$.
    \item The Gumbel-Softmax temperature $\tau $ is fixed at 1 throughout the training procedure.
\end{itemize}

\textbf{Evaluation Protocol and Decoding Configuration}
\begin{itemize}
    \item All zero-shot benchmark evaluations are conducted using the OpenCompass~\cite{2023opencompass} framework in generation mode.
    \item Prompting is standardized by utilizing the official Transformers chat templates for the respective models.
    \item For the Qwen3 model family, the model-specific "thinking" mode is explicitly disabled to ensure comparable outputs across all baselines.
    \item A consistent decoding configuration is applied across all core benchmarks (ARC-e, ARC-c, PIQA, HellaSwag, WinoGrande, and BoolQ): temperature = 0.7, top\_p = 0.8, top\_k = 20, and min\_p = 0.
\end{itemize}

\textbf{Hardware and Inference Efficiency Measurements}
\begin{itemize}
    \item The preference score learning and subsequent inference throughput evaluations are executed on NVIDIA A100 GPUs. For the Qwen3-8B model, the Stage I score optimization requires approximately 1 hour, whereas the execution time for the Stage II integer-programming projection is negligible.
    \item Inference throughput is evaluated using the standard AQLM benchmarking utility (generate\_benchmark.py), backed by Triton kernels, with a batch size of 1 and an output generation length of 128 tokens, reporting the average performance over 10 consecutive trials.
\end{itemize}

\section{Heatmaps across budgets and calibration sets}
\label{app:heatmaps}

\begin{figure*}[htbp] 
     \centering
     \begin{subfigure}[b]{0.23\textwidth}
         \centering
         \includegraphics[width=\textwidth]{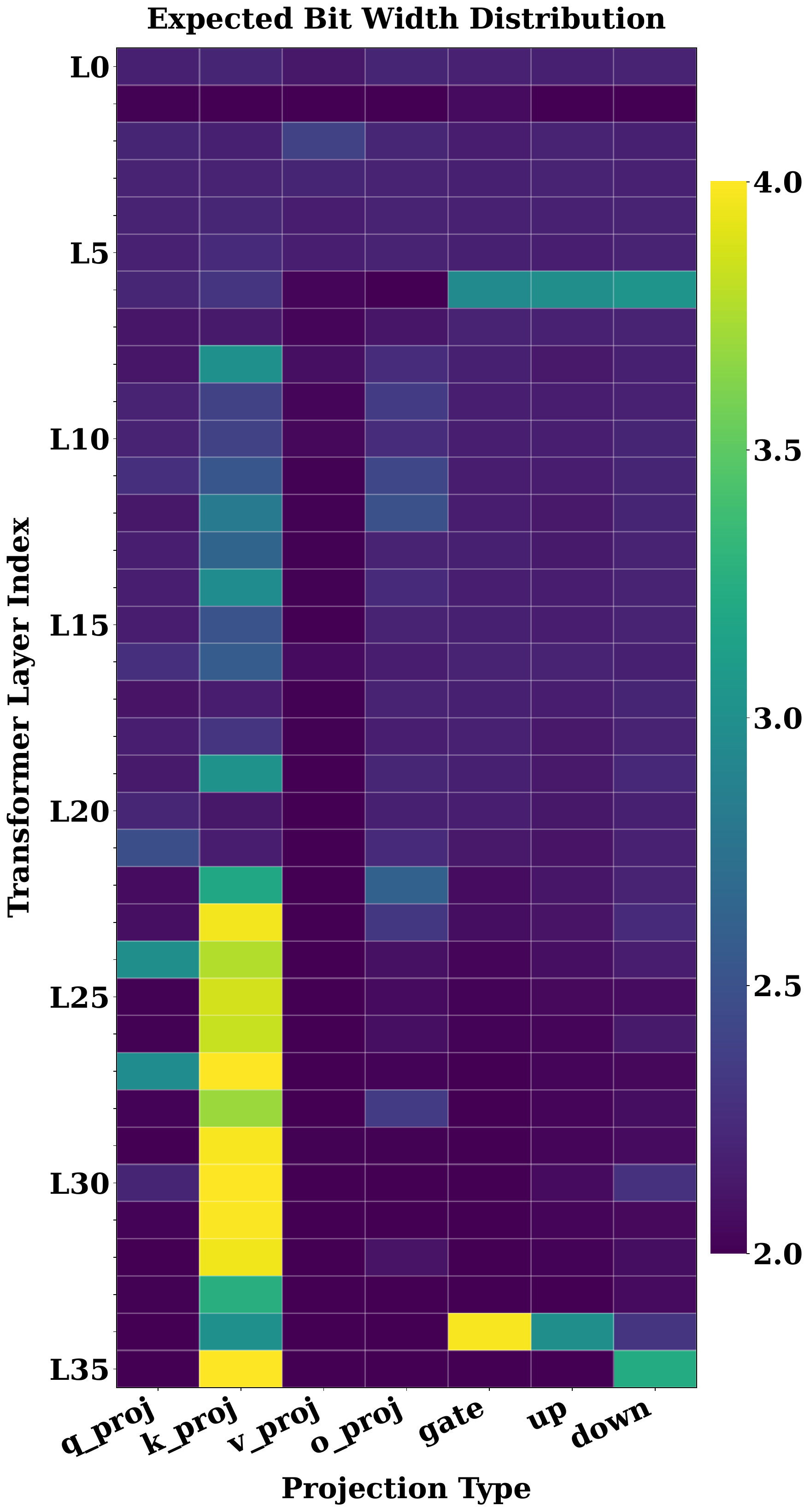}
         \caption{RedPajama $b_{\text{target}}{=}2.2$}
         \label{app:pajama_22}
     \end{subfigure}
     \hfill
     \begin{subfigure}[b]{0.23\textwidth}
         \centering
          \includegraphics[width=\textwidth]
{figures/refined_bitwidth_heatmap_pajama_2.5.pdf}
         \caption{RedPajama $b_{\text{target}}{=}2.5$}
         \label{app:pajama_25}
     \end{subfigure}
     \hfill
     \begin{subfigure}[b]{0.23\textwidth}
         \centering
          \includegraphics[width=\textwidth]
{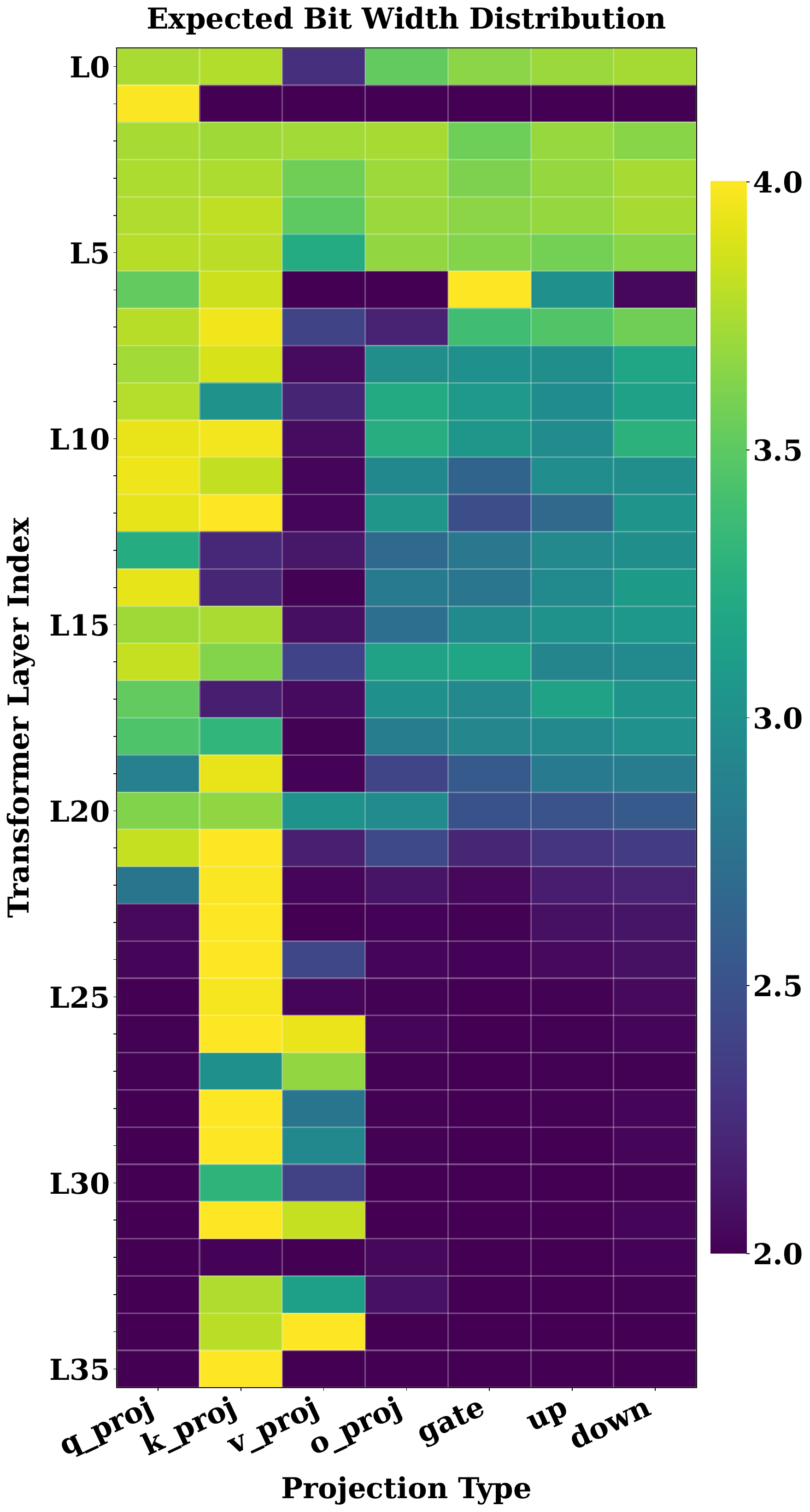}
         \caption{RedPajama $b_{\text{target}}{=}2.7$}
         \label{app:pajama_27}
     \end{subfigure}
          \hfill
     \begin{subfigure}[b]{0.23\textwidth}
         \centering
          \includegraphics[width=\textwidth]
{figures/refined_bitwidth_heatmap_pajama_3.0.pdf}
         \caption{RedPajama $b_{\text{target}}{=}3.0$}
         \label{app:pajama_30}
     \end{subfigure}

     \vspace{8pt} 

     \begin{subfigure}[b]{0.23\textwidth}
         \centering
          \includegraphics[width=\textwidth]
{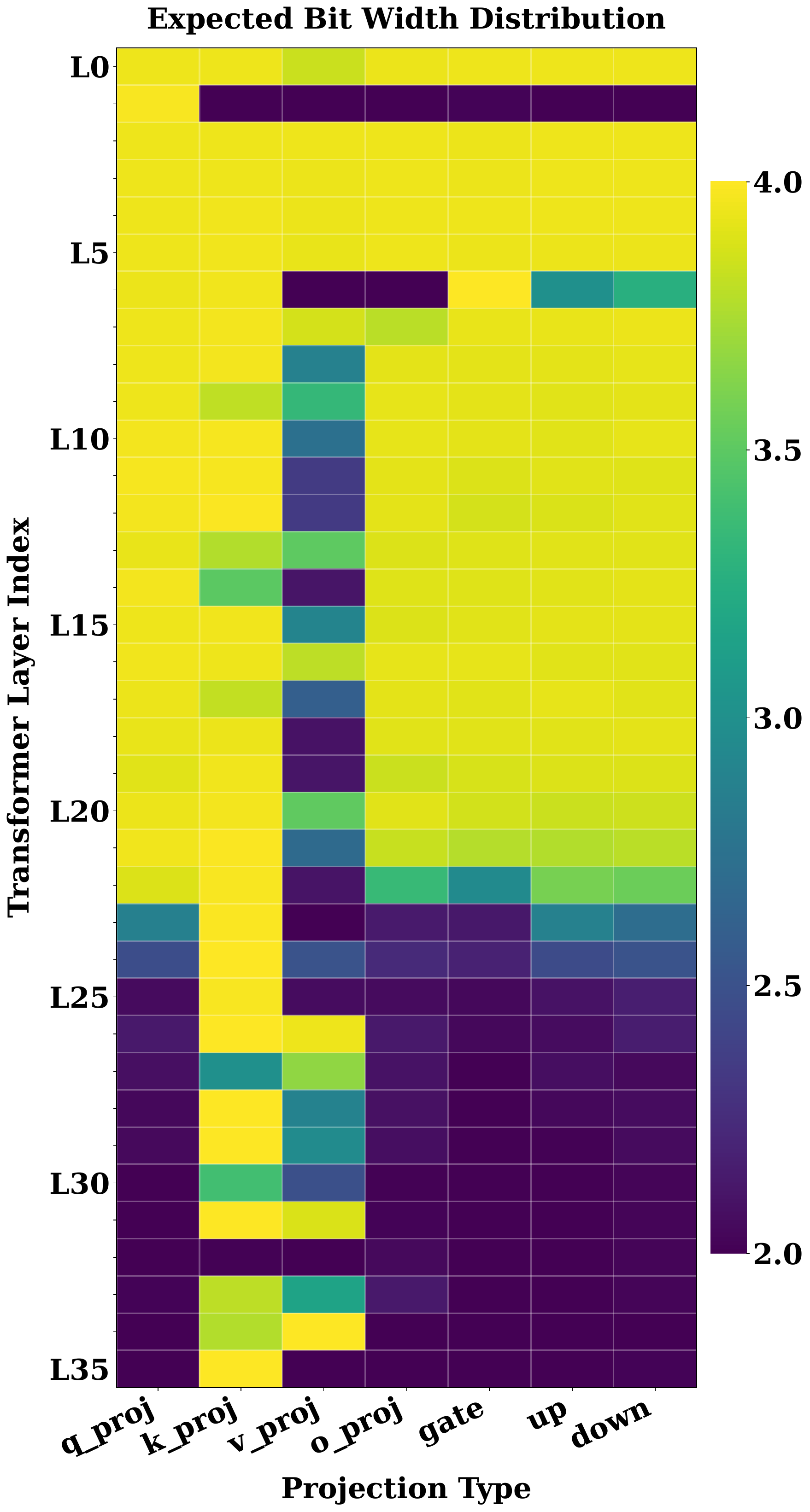}
         \caption{RedPajama $b_{\text{target}}{=}3.2$}
         \label{app:pajama_32}
     \end{subfigure}
     \hfill
     \begin{subfigure}[b]{0.23\textwidth}
         \centering
          \includegraphics[width=\textwidth]
{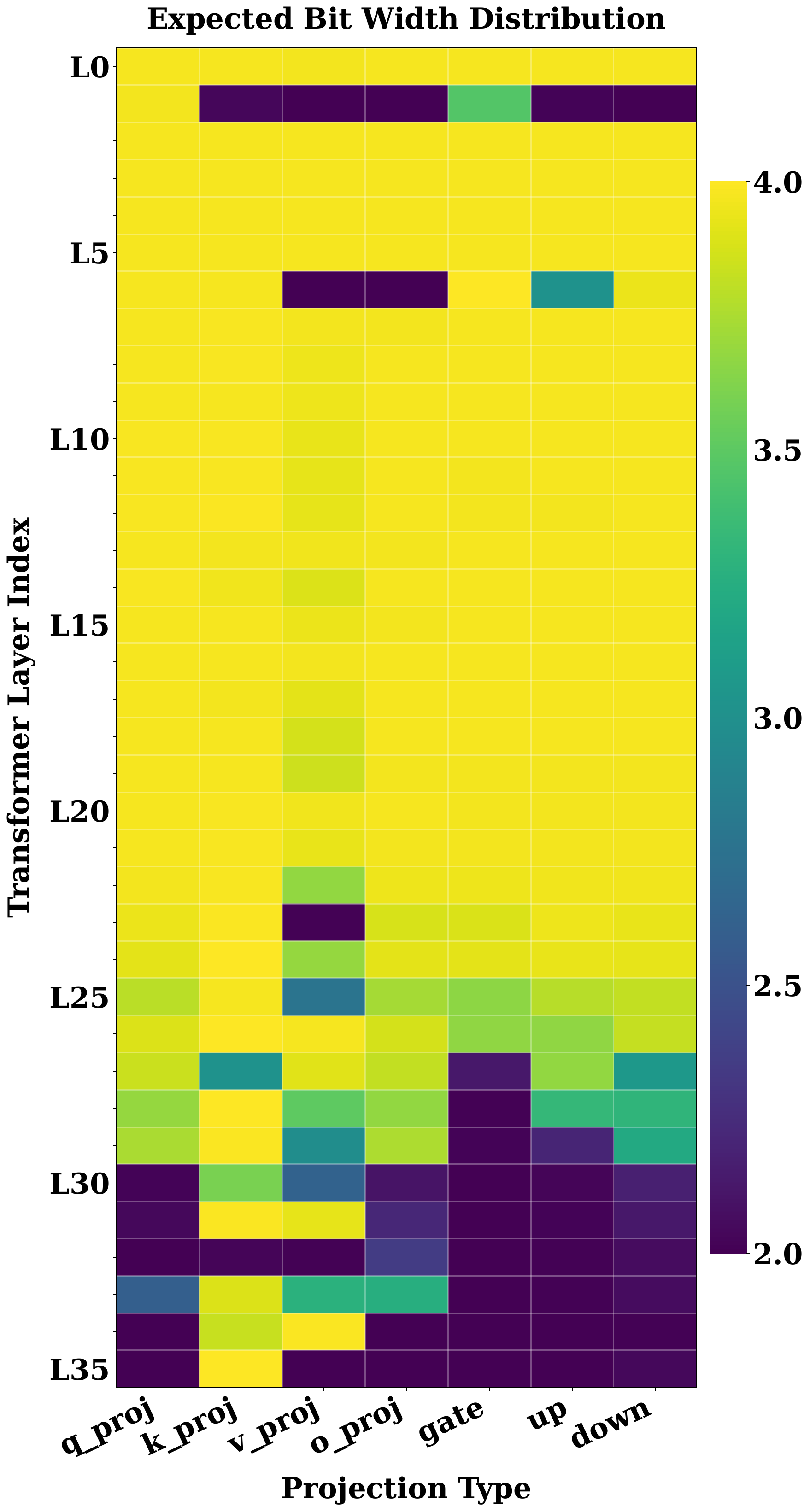}
         \caption{RedPajama $b_{\text{target}}{=}3.5$}
         \label{app:pajama_35}
     \end{subfigure}
     \hfill
     \begin{subfigure}[b]{0.23\textwidth}
         \centering
          \includegraphics[width=\textwidth]
{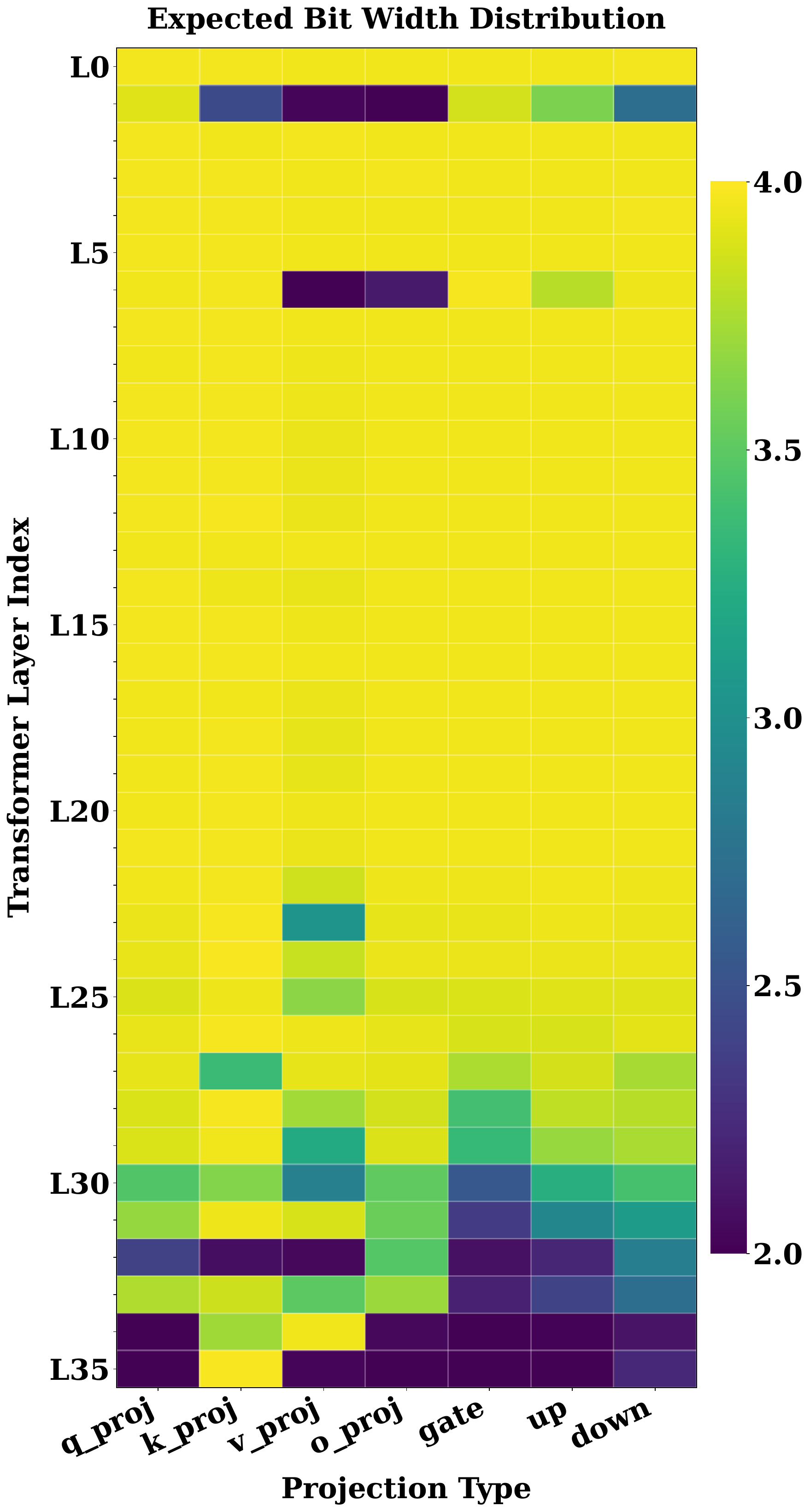}
         \caption{RedPajama $b_{\text{target}}{=}3.7$}
         \label{app:pajama_37}
     \end{subfigure}
         \hfill
     \begin{subfigure}[b]{0.23\textwidth}
         \centering
          \includegraphics[width=\textwidth]
            {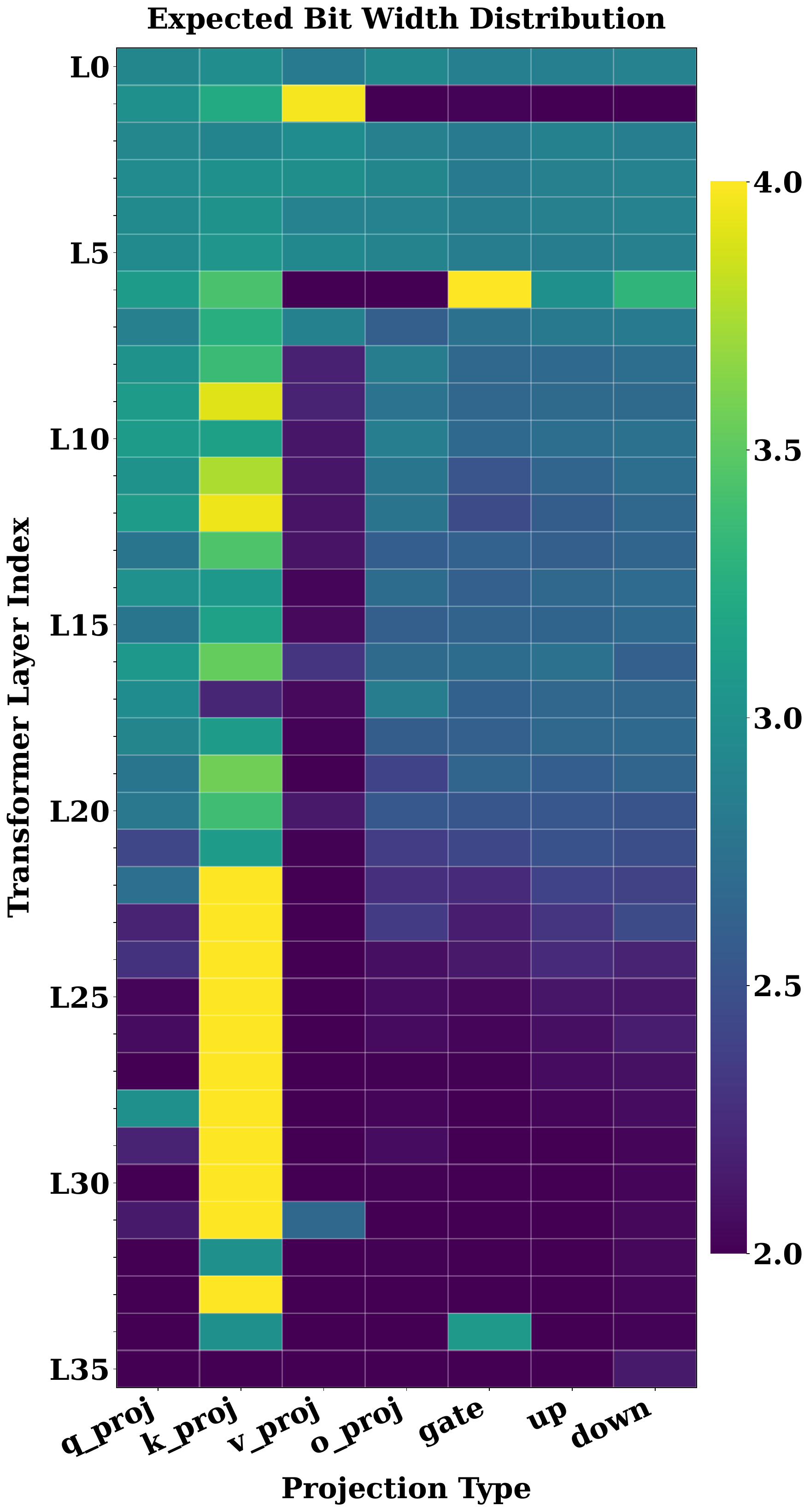}
         \caption{WikiText $b_{\text{target}}{=}2.5$}
         \label{app:wiki_25}
     \end{subfigure}
     
\caption{\textbf{Additional bit-width heatmaps across budgets and calibration sets.}
Expected bit-width assigned by \MethodName~ to each projection type (x-axis) at every Transformer layer (y-axis).
Panels (a)--(g) vary the target average bit-width $b_{\text{target}}$ on the RedPajama calibration set, showing a consistent allocation structure as the budget increases.
Panel (h) reports the same visualization on WikiText at $b_{\text{target}}{=}2.5$, illustrating that the learned pattern is largely consistent across calibration distributions.}
\label{fig:heatmaps_app}
\end{figure*}

\end{document}